\documentclass[10pt,twocolumn,letterpaper]{article}

\usepackage{cvpr}
\usepackage{times}
\usepackage{epsfig}
\usepackage{graphicx}
\usepackage{amsmath}
\usepackage{amssymb}

\usepackage{algorithmic}
\usepackage{makecell}
\usepackage{lineno}
\usepackage{graphicx}
\usepackage{mathrsfs}
\usepackage{algorithm}
\usepackage{bm}

\usepackage{authblk}

\usepackage[pagebackref=true,breaklinks=true,letterpaper=true,colorlinks,bookmarks=false]{hyperref}

\cvprfinalcopy 

\def\cvprPaperID{2624} 

\ifcvprfinal\fi
\begin{document}

\title{Towards Universal Representation for Unseen Action Recognition}

	\author[1]{Yi Zhu}
	\author[2]{Yang Long\thanks{Yang Long contributed equally to this work.}}
	\author[2]{Yu Guan}
	\author[1]{Shawn Newsam}
	\author[3]{Ling Shao}
	
	\affil[1]{University of California, Merced}
	\affil[2]{Open Lab, School of Computing, Newcastle University, UK.}
	\affil[3]{Inception Institute of Artificial Intelligence (IIAI), Abu Dhabi, UAE.}
	
	

\maketitle
\thispagestyle{empty}

\begin{abstract}
	Unseen Action Recognition (UAR) aims to recognise novel action categories without training examples. While previous methods focus on inner-dataset seen/unseen splits, this paper proposes a pipeline using a large-scale training source to achieve a \textit{Universal Representation} (UR) that can generalise to a more realistic Cross-Dataset UAR (CD-UAR) scenario. We first address UAR as a \textit{Generalised Multiple-Instance Learning} (GMIL) problem and discover `building-blocks' from the large-scale ActivityNet dataset using distribution kernels. Essential visual and semantic components are preserved in a shared space to achieve the UR that can efficiently generalise to new datasets. Predicted UR exemplars can be improved by a simple semantic adaptation, and then an unseen action can be directly recognised using UR during the test. Without further training, extensive experiments manifest significant improvements over the UCF101 and HMDB51 benchmarks.
\end{abstract}

\section{Introduction}

The field of human action recognition has advanced rapidly over the past few years. We have moved from manually designed features \cite{idtfWang2013,videoDarwin} to learned convolutional neural network (CNN) features \cite{c3d2015,KarpathyCVPR14}; from encoding appearance information to encoding motion information \cite{twostream2014,longTemporalConv2016,atf_cvpr17}; and from learning local features to learning global video features \cite{TSN2016ECCV, diba_tle_2016,dovf_lan_2017}. The performance has continued to soar higher as we incorporate more of the steps into an end-to-end learning framework \cite{depth2action,hidden_zhu_17}. However, such robust and accurate action classifiers often rely on large-scale training video datasets using deep neural networks, which require large numbers of expensive annotated samples per action class. Although several large-scale video datasets have been proposed like Sports-1M \cite{KarpathyCVPR14}, ActivityNet \cite{activityNet}, YouTube-8M \cite{YouTube_8M_2016} and Kinetics \cite{kinetics}, it is practically infeasible and extremely costly to annotate action videos with the ever-growing need of new categories. 

Zero-shot action recognition has recently drawn considerable attention because of its ability to recognize unseen action categories without any labelled examples. The key idea is to make a trained model that can generalise to unseen categories with a shared semantic representation. The most popular side information being used are attributes, word vectors and visual-semantic embeddings. Such zero-shot learning frameworks effectively bypass the data collection limitations of traditional supervised learning approaches, which makes them more promising paradigms for UAR.

\begin{figure*}
	\centering
	\includegraphics[width=0.85\textwidth]{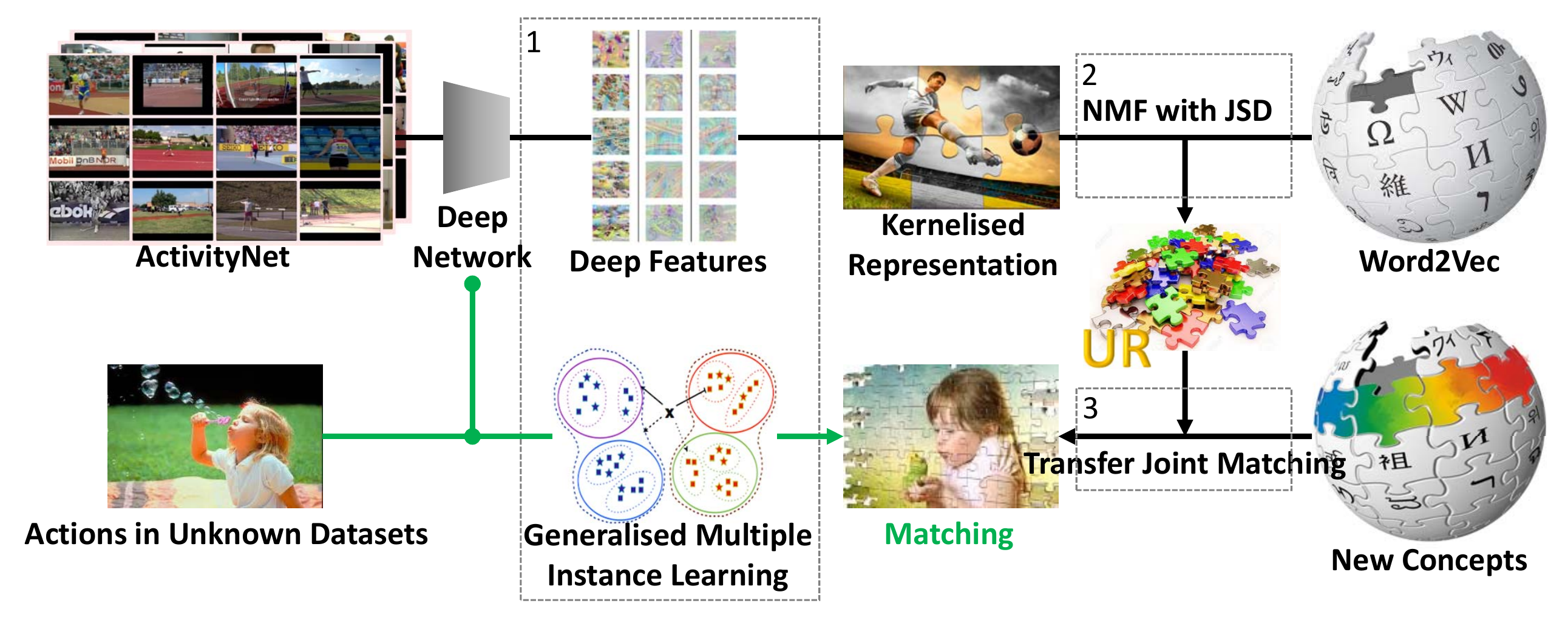}
	\vspace{-1ex}
	\caption{The proposed CD-UAR pipeline: 1) Extract deep features for each frame and summarise the video by essential components that are kernelised by GMIL; 2) Preserve shared components with the label embedding to achieve UR using NMF with JSD; 3) New concepts can be represented by UR and adjusted by domain adaptation. Test (green line): unseen actions are encoded by GMIL using the same essential components in ActivityNet to achieve a matching using UR. \label{framework}}
	\vspace{-2ex}
\end{figure*}

Extensive work on zero-shot action recognition has been done in the past five years. \cite{2011_Liu_zsl_action,latent_attr_Fu_PAMI13,mutlisource_zs_Gan_CVPR16,myACMMM,my_2017_PAMI} considered using attributes for classifications. These attribute-based methods are easy to understand and implement, but hard to define and scale up to a large-scale scenario. Semantic representations like word vectors \cite{tme_zs_Fu_ECCV14,synonyms_Alexiou_ICIP2016,myAAAI_18} are thus preferred since only category names are required for constructing the label embeddings. There also has been much recent work on using visual-semantic embeddings extracted from pre-trained deep networks \cite{2015_ICCV_ZSL_detection,object_scene_Wu_CVPR16,2017_ICCV_Spatial_aware} due to their superior performance over single view word vectors or attributes. 

However, whichever side information we adopt, the generalisation capability of these approaches is not promising, which is referred to as the domain shift problem. Most previous work thus still focuses on inner-dataset seen/unseen splits. This is not very practical since each new dataset or each category will require re-training. Motivated by such a fact we propose to utilise a large-scale training source to achieve a \textit{Universal Representation} (UR) that can automatically generalise to a more realistic Cross-Dataset UAR (CD-UAR) scenario. Unseen actions from new datasets can be directly recognised via the UR without further training or fine-tuning on the target dataset. 

The proposed pipeline is illustrated in Fig. \ref{framework}. We first leverage the power of deep neural networks to extract visual features, which results in a Generative Multiple Instance Learning (GMIL) problem. Namely, all the visual features (instances) in a video share the label while only a small portion is determinative. Compared to conventional global summaries of visual features using Bag-of-Visual-Word or Fisher Vector encoding, GMIL aims to discover those essential ``building-blocks'' to represent actions in the source and target domains and suppress the ambiguous instances. We then introduce our novel Universal Representation Learning (URL) algorithm composed of Non-negative Matrix Factorisation (NMF) with a Jensen$\text{-}$Shannon Divergence (JSD) constraint. The non-negativity property of NMF allows us to learn a part-based representation, which serves as the key bases between the visual and semantic modalities. JSD is a symmetrised and bounded version of the Kullback-Leibler divergence, which can make balanced generalisation to new distributions of both visual and semantic features. A representation that can generalise to both visual and semantic views, and both source and target domains, is referred to as the UR. More insighs of NMF, JSD, and UR will be discussed in the experiments. Our main contributions can be summarised as follows:
\begin{itemize}
	\item This paper extends conventional UAR tasks to more realistic CD-UAR scenarios. Unseen actions in new datasets can be directly recognised via the UR without further training or fine-tuning on the target dataset.
	\item We propose a CD-UAR pipeline that incorporates deep feature extraction, Generative Multiple Instance Learning, Universal Representation Learning, and semantic domain adaptation.
	\item Our novel URL algorithm unifies NMF with a JSD constraint. The resultant UR can substantially preserve both the shared and generative bases of visual semantic features so as to withstand the challenging CD-UAR scenario.
	\item Extensive experiments manifest that the UR can effectively generalise across different datasets and outperform state-of-the-art approaches in inductive UAR scenarios using either low-level or deep features.
\end{itemize}

\section{Related Work}

Zero-shot human action recognition has advanced rapidly due to its importance and necessity as aforementioned. The common practice of zero-shot learning is to transfer action knowledge through a semantic embedding space, such as attributes, word vectors or visual features.

Initial work \cite{2011_Liu_zsl_action} has considered a set of manually defined attributes to describe the spatial-temporal evolution of the action in a video. 
Gan et al. \cite{mutlisource_zs_Gan_CVPR16} investigated the problem of how to accurately and robustly detect attributes from images or videos, and the learned high-quality attribute detectors are shown to generalize well across different categories. However, attribute-based methods suffer from several drawbacks: (1) Actions are complex compositions including various human motions and human-object interaction. It is extremely hard (e.g., subjective, labor-intensive, lack of domain knowledge) to determine a set of attributes for describing all actions; (2) Attribute-based approaches are not applicable for large-scale settings since they always require re-training of the model when adding new attributes; (3) Despite the fact that the attributes can be data-driven learned or semi-automatically defined \cite{latent_attr_Fu_PAMI13}, their semantic meanings may be unknown or inappropriate. 

Hence, word vectors have been preferred for zero-shot action recognition, since only category names are required for constructing the label embeddings.
\cite{tme_zs_Fu_ECCV14,2015_ICIP_Xuxun} are among the first works to adopt semantic word vector spaces as the intermediate-level embedding for zero-shot action recognition. Following \cite{2015_ICIP_Xuxun}, Alexiou et al. \cite{synonyms_Alexiou_ICIP2016} proposed to explore broader semantic contextual information (e.g., synonyms) in the text domain to enrich the word vector representation of action classes. However, word vectors alone are deficient for discriminating various classes because of the semantic gap between visual and textual information. 

Thus, a large number of recent works \cite{2015_ICCV_ZSL_detection,domain_adapt_Li_ICIP16,object_scene_Wu_CVPR16,alter_semantic_Wang_PKDD17} exploit large object/scene recognition datasets to map object/scene scores in videos to actions. This makes sense since objects and scenes could serve as the basis to construct arbitrary action videos and the semantic representation can alleviate such visual gaps. The motivation can also be ascribed to the success of CNNs \cite{material_jiaxue_cvpr17,idr_CGAN_zhang17,ground_jiaxue_cvpr18}. With the help of off-the-shelf object detectors, such methods \cite{2017_ICCV_Spatial_aware} could even perform zero-shot spatio-temporal action localization. 

There are also other alternatives to solve zero-shot action recognition. Gan et al. \cite{SIR_Gan_AAAI15} leveraged the semantic inter-class relationships between the known and unknown actions followed by label transfer learning. Such similarity mapping doesn't require attributes. Qin et al. \cite{2017_CVPR_errorCorection} formulated zero-shot learning as designing error-correcting output codes, which bypass the drawbacks of using attributes or word vectors. Due to the domain shift problem, several works have extended the methods above using either transductive learning \cite{tme_zs_Fu_ECCV14,tran_zsar_Xu_IJCV17} or domain adaptation \cite{unsup_domain_Kodirov_ICCV15,2016_XuXun_prioristised}.

However, all previous methods focus on inner-dataset seen/unseen splits while we extend the problem to CD-UAR. This scenario is more realistic and practical; for example, we can directly recognise unseen categories from new datasets without further training or fine-tuning. Though promising, CD-UAR is much more challenging compared to conventional UAR. 
We contend that when both CD and UAR are considered, the severe domain shift exceeds the generalization capability of existing approaches. Hence, we propose the URL algorithm to obtain a more robust universal representation. Our novel CD-UAR pipeline dramatically outperforms both conventional benchmarks and state-of-the-art approaches, which are in inductive UAR scenarios using low-level features and CD-UAR using deep features, respectively. One related work also applies NMF to zero-shot image classification \cite{2017_CVPR_NMF}. Despite the fact that promising generalisation is reported, which supports our insights, it still focuses on inner-class splits without considering CD-UAR. Also, their sparsity constrained NMF has completely different goals to our methods with JSD.

\section{Approach}
In this section, we first formalise the problem and clarify each step as below. We then introduce our CD-UAR pipeline in detail, which includes Genearalised Multiple-Instance Learning, Universal Representation Learning and semantic adaptation. 

\noindent\textbf{Training} 
Let $(\bm{x}_1,y_1),\cdots,(\bm{x}_{N_s},y_{N_s})\subseteq\bm{X}_s\times\bm{Y}_s$ denote the training actions and their class labels in pairs in the source domain $\mathcal{D}_s$, where $N_s$ is the training sample size; each action $\bm{x}_{i}$ has $L_i$ frames in a $D$-dimensional visual feature space $[\bm{x}_{i}]=(\bm{x}_{i}^1,...,\bm{x}_{i}^{L_i})\in \mathbb{R}^{D\times L_i}$; $y_i\in\{1,\cdots,C\} $ consists of $C$ discrete labels of training classes. 

\noindent\textbf{Inference} Given a new dataset in the target domain $\mathcal{D}_t$ with $C_u$ unseen action classes that are novel and distinct, \ie $\bm{Y}_u=\{C+1,...,C+C_u\}$ and $\bm{Y}_u\cap\bm{Y}_s=\emptyset$, the key solution to UAR needs to associate these novel concepts to $\mathcal{D}_s$ by human teaching. To avoid expensive annotations, we adopt Word2vec semantic ($\bm{S}$) label embedding $(\bm{\hat{s}}_1, \hat{y}_1),\cdots, (\bm{\hat{s}}_{C_u}, \hat{y}_{C_u}) \subseteq \bm{S}_u \times \bm{Y}_u$. Hat and subscript $u$ denote information about unseen classes. Inference then can be  achieved by learning a visual-semantic compatibility function $\min\mathcal{L}(\Phi(\bm{X}_s),\Psi(\bm{S}_s))$ that can generalise to $\bm{S}_u$.

\noindent\textbf{Test} Using the learned $\mathcal{L}$, an unseen action $\hat{x}$ can be recognised by $f: \Phi(\hat{\bm{x}}) \rightarrow \Psi(\bm{S}_u)\times \bm{Y}_u$.

\begin{figure}
	\centering
	\includegraphics[width=0.35\textwidth]{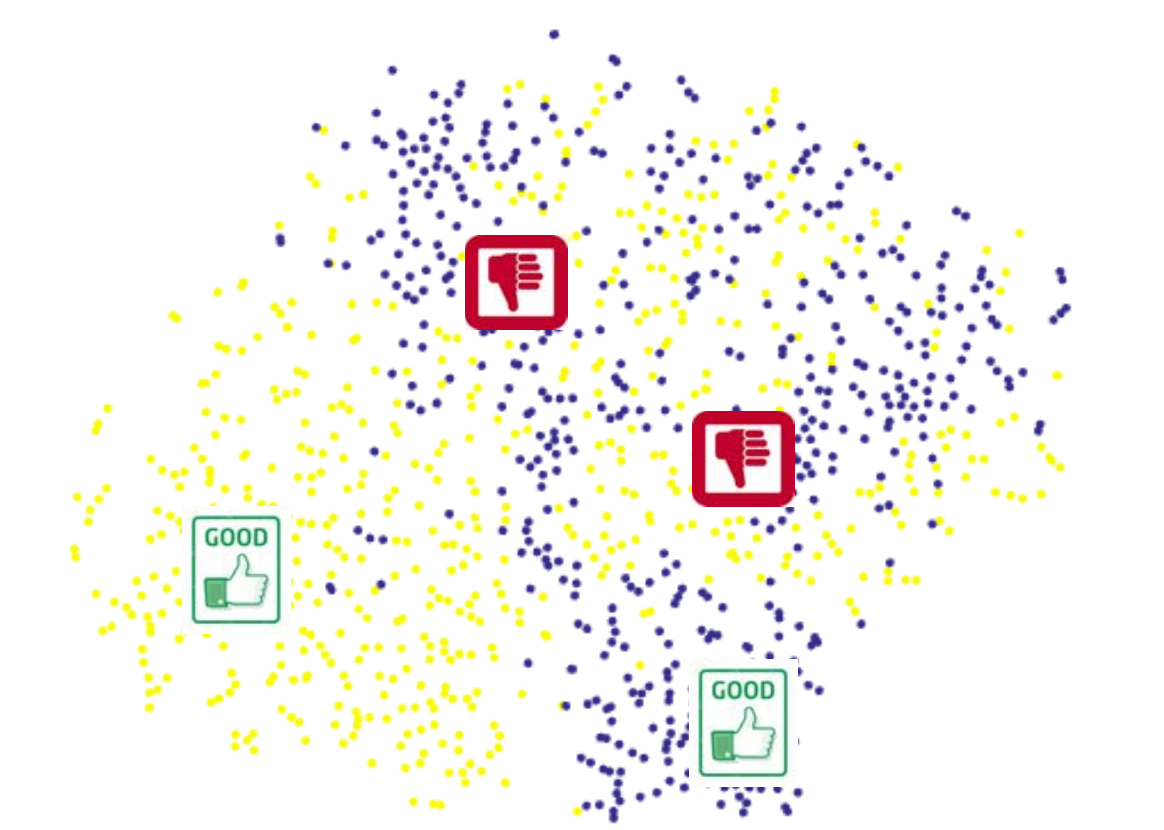}
	\vspace{1ex}
	\caption{Visualisation of feature distributions of action `long-jump' and `triple-jump' in the ActivityNet dataset using tSNE. \label{tSNE}}
	\vspace{-2ex}
\end{figure}

\subsection{Genearalised Multiple-Instance Learning}
Conventional summary of $\bm{x}_i$ can be achieved by Bag-of-Visual-Words or Fisher Vectors \cite{FisherVector}. 
In GMIL, it is assumed that instances in the same class can be drawn from different distributions. Let $P(\cdot)$ denote the space of Borel probability measures over its argument, which is known as a bag. Conventionally, it is assumed that some instances are \textit{attractive} $P_+(\bm{x})$ while others are \textit{repulsive} $P_-(\bm{x})$. This paper argues that many instances may exist in \textit{neutral} bags. In Fig. \ref{tSNE}, we show an example of visual feature distributions of `long-jump' and `triple-jump'. Each point denotes a frame. While most frames fall in the neutral bags (red thumb), only a few frames (green thumb) are attractive to one class and repulsive to others. The neutral bags may contain many basic action bases shared by classes or just background noise. Conventional \textit{Maximum Mean Discrepancy} \cite{GMIL_2016_IJCAI} may not well represent such distributions. Instead, this paper adopts the odds ratio embedding, which aims to discover the most \textit{attractive} bases to each class $c$ and suppress the neutral ones. This can be simply implemented by the pooled Naive Bayes Nearest Neighbor (NBNN) kernel \cite{pooled_nbnn} at the `bag-level'. We conduct $k$-means on each class to cluster them into $H$ bags. The associated kernel function is:
\begin{equation}
k(\bm{x},\bm{x}')=\phi(\bm{x})^T\phi(\bm{x}'),
\label{eq_nbnn_kernel}
\end{equation}
where $\Phi(\bm{X}_s)=[\phi(\bm{x}_i)]=[\phi^1(\bm{x}_i),...,\phi^{CH}(\bm{x}_i)]^T$ is the kernelised representation with odds ratio \cite{local_nbnn} applies to each kernel embedding: $
\phi^{c_i}(\bm{x})=\sum_{l=1}^{L_i}\log\frac{p(c|\bm{x}^l)}{p(\bar{c}|\bm{x}^l)}$. Specific implementation details can be found in the supplementary material. In this way, we discover $C\times H$ bases as `building-blocks' to represent any actions in both the source and target domains.

\vspace{-1ex}
\subsection{Universal Representation Learning}
For clarity, we use $A=\Phi(\bm{X}_s)$ and $B=\bm{S}_s$ to define the visual and semantic embeddings in the source domain $\mathcal{D}_{s}:A\times B$. Towards universal representation, we aim to find a shared space that can: 1) well preserve the key bases between visual and semantic modalities; 2) generalise to new distributions of unseen datasets. 
For the former, let $A=[\mathbf{a}_{1}, \cdots, \mathbf{a}_{N_{s}}]\in \mathbb{R}^{M_{1}\times N_{s}}_{\ge 0}$ and $B=[\mathbf{b}_{1}, \cdots, \mathbf{b}_{N_{s}}]\in \mathbb{R}^{M_{2}\times N_{s}}_{\ge 0}$; $M_1=C\times H$ and $M_{2}=L$. NMF is employed to find two nonnegative matrices from $A$: $U \in \mathbb{R}^{M_{1}\times D_{1}}_{\ge 0}$ and $V_{1} \in \mathbb{R}^{D_{1}\times N_{s}}_{\ge 0}$ and two nonnegative matrices from $B$: $W \in \mathbb{R}^{M_{2}\times D_{2}}_{\ge 0}$ and $V_{2} \in \mathbb{R}^{D_{2}\times N_{s}}_{\ge 0}$ with full rank whose product can approximately represent the original matrix $A$ and $B$, \emph{i.e.}, $A\approx UV_{1}$ and $B\approx WV_{2}$. In practice, we set $D_{1}<$ min$(M_{1}, N_{s})$ and $D_{2}<$ min$(M_{2}, N_{s})$. 
We constrain the shared coefficient matrix: $V_{1}=V_{2}=V \in \mathbb{R}^{D\times N_{s}}_{\ge 0}$. 
For the latter aim, we introduce JSD to preserve the generative components from the GMIL and use these essential `building-blocks' to generalise to unseen datasets. Hence, the overall objective function is given as:
\begin{equation}\label{eq_OBJ}
\begin{split}
\mathcal{L}=\mathop{\textup{min}}\limits_{U,W,V}  & \|A-UV\|_F^2\!+\! \|B-WV\|_F^2 \! \\
&+ \!\eta \  \text{JSD}, s.t. \   U, W, V \ge 0,
\end{split}
\end{equation}
where $\| \!\cdot\! \|_F$ is the Frobenius norm; $\eta$ is a smoothness parameter; $\text{JSD}$ is short for the following equation:
\begin{equation}\label{eq_JSD}
\begin{split}
\text{JSD}(P_{A}||P_{B})&=\frac{1}{2}KL(P_{A}||Q)+\frac{1}{2}KL(P_{B}||Q)\\
&= \frac{1}{2} \sum_{i}\sum_{j}p_{A}^{ij}\log{p_{A}^{ij}}-p_{A}^{ij}\log{q_{ij}} \\
& + \frac{1}{2} \sum_{i}\sum_{j}p_{B}^{ij}\log{p_{B}^{ij}}-p_{B}^{ij}\log{q_{ij}}
\end{split}
\end{equation}
where $P_{A}$ and $P_{B}$ are probability distributions in space $A$ and $B$. We aim to find the joint probability distribution $Q$ in the shared space $V$ that is generalised to by $P_{A}$ and $P_{B}$ and their shifted distributions in the target domain. Specifically, $\text{JSD}$ can be estimated pairwise as:
\begin{equation}\label{eq_pa}
\left\lbrace
\begin{split}
p_{A}^{ij}&=  \frac{g(\mathbf{a}^i,\mathbf{a}^j)}{\sum_{k \neq l} g(\mathbf{a}^k,\mathbf{a}^l)}\\
p_{B}^{ij}&=  \frac{g(\mathbf{b}^i,\mathbf{b}^j)}{\sum_{k \neq l} g(\mathbf{b}^k,\mathbf{b}^l)}\\
q_{ij}&= \frac{(1+\|\mathbf{v}_i-\mathbf{v}_j\|^2)^{-1}}{\sum_{k \neq l} (1+ \|\mathbf{v}_k - \mathbf{v}_l \|^2)^{-1}}.
\end{split}
\right.
\end{equation}

Without loss of generality, this paper use the cross-entropy distance to implement $g(\cdot)$.
\vspace{-1ex}
\subsubsection{Optimization}
Let the Lagrangian of Eq. \ref{eq_OBJ} be:
\begin{equation}\label{}
	\begin{split}
		\mathcal{L} &=\|A-UV \|^2+ \|B-WV\|^2 + \eta \  \text{JSD}  \\
		&+ tr(\Phi U^T) + tr(\Theta W^T)+ tr (\Psi V^T),
	\end{split}
\end{equation}
where $\Phi$, $\Theta$ and $\Psi$ are three Lagrangian multiplier matrices. $tr(\cdot)$ denotes the trace of a matrix. For clarity, $\text{JSD}$ in Eq. \ref{eq_JSD} is simply denoted as $G$. We define two auxiliary variables $d_{ij}$ and $Z$ as follows:
\begin{equation}
	d_{ij}=\|\mathbf{v}_i-\mathbf{v}_j\| ~\text{and}~ Z={\sum_{k \neq l} (1+ d_{kl}^2)^{-1}}.
\end{equation}

Note that if $\mathbf{v}_i$ changes, the only pairwise distances that change are $d_{ij}$ and $d_{ji}$. Therefore, the gradient of function $G$ with respect to $\mathbf{v}_i$ is given by:
\begin{equation}\label{gvder}
	\frac{\partial G}{\partial \mathbf{v}_i} = 2\sum_{j=1}^N \frac{\partial G}{\partial {d}_{ij}}(\mathbf{v}_i-\mathbf{v}_j).
\end{equation}

Then $\frac{\partial G}{\partial {d}_{ij}}$ can be calculated by JS divergence in Eq. (\ref{eq_JSD}):
\begin{equation}
\resizebox{0.43\textwidth}{!}{$
	\begin{split}
		\frac{\partial G}{\partial {d}_{ij}} \!\!=\!\!-\!\frac{\eta}{2}\! \sum_{k \!\neq \! l}(p_{A}^{kl}\!\!+\!\!p_{B}^{kl}\!)\left(\!\frac{1}{q_{kl}Z}\frac{\partial ((1\!\!+ \!\! d_{kl}^2)^{-1})}{\partial {d}_{ij}}\!\!-\!\!\frac{1}{Z}\frac{\partial Z}{\partial {d}_{ij}} \right) .
	\end{split}
	$}
\end{equation}

Since $\frac{\partial ((1+ d_{kl}^2)^{-1})}{\partial {d}_{ij}}$ is nonzero if and only if $k=i$ and $l=j$, and $\sum_{k \neq l}p_{kl}=1$, it can be simplified as:
\begin{equation}\label{gdder}
	\frac{\partial G}{\partial {d}_{ij}}=\eta (p_{A}^{ij}+p_{B}^{ij}-2q_{ij})(1+d_{ij}^2)^{-1}.
\end{equation}

Substituting Eq. (\ref{gdder}) into Eq. (\ref{gvder}), we have the gradient of the JS divergence as:
\begin{equation}\label{G}
	\begin{split}
		\frac{\partial G}{\partial \mathbf{v}_i}\! \!=\!\! 2\eta \sum_{j\!=\!1}^N (p_{A}^{ij}\!\!+\!\!p_{B}^{ij}\!\!-\!\!2q_{ij}) (\mathbf{v}_i \!\!- \!\!\mathbf{v}_j) (1\!\!+\!\! \|\mathbf{v}_i\!\!-\!\! \mathbf{v}_j\|^2)^{\!-\!1}.
	\end{split}
\end{equation}

Let the gradients of $\mathcal{L}$ be zeros to minimize $O_f$:
\begin{equation}\label{1}
	\frac{\partial \mathcal{L}}{\partial V}\!\! =\!\! 2(-U^TA \!\!+\!\! U^TUV \!\!-\!\! W^TB \!\!+\!\! W^TWV)\!\!+\!\! \frac{\partial G}{\partial V}
	\!\!+\!\! \Psi \!\! =\!\! \mathbf{0},
\end{equation}

\begin{equation}\label{2}
	\frac{\partial \mathcal{L}}{\partial U} = 2(-AV^T + UVV^T) + \Phi = \mathbf{0},
\end{equation}

\begin{equation}\label{3}
	\frac{\partial \mathcal{L}}{\partial W} = 2(-BW^T + WVV^T) + \Theta = \mathbf{0}.
\end{equation}

In addition, we also have KKT conditions: $\Phi_{ij} U_{ij} = 0$, $\Theta_{ij} W_{ij} = 0$ and $\Psi_{ij} V_{ij} = 0$, $\forall i,j$. Then multiplying $V_{ij}$, $U_{ij}$ and $W_{ij}$ in the corresponding positions on both sides of Eqs. (\ref{1}), (\ref{2}) and (\ref{3}) respectively, we obtain:
\begin{equation}\label{4}
\resizebox{0.42\textwidth}{!}{$
	\left(2(-U^TA \!\!+\!\! U^TUV \!\!-\!\!W^TB \!\!+\!\! W^TWV) \!\!+\!\! \frac{\partial G}{\partial \mathbf{v}_i}\right)_{ij} V_{ij}\!\! = \!\!0,
	$}
\end{equation}
\begin{equation}\label{5}
	2(-AV^T + UVV^T)_{ij} U_{ij} = 0,
\end{equation}
\begin{equation}\label{6}
	2(-BV^T + WVV^T)_{ij} W_{ij} = 0.
\end{equation}

Note that
\begin{equation*}
\resizebox{0.42\textwidth}{!}{$
	\begin{split}
		\left(\frac{\partial G}{\partial \mathbf{v}_j}\right)_i &= \left(2\eta \sum_{k=1}^N \frac{(p_{A}^{jk} + p_{B}^{jk} -2q_{jk})(\mathbf{v}_j-\mathbf{v}_k)} {1+\|\mathbf{v}_j-\mathbf{v}_k\|^2}\right)_i \\
		&= 2\eta \sum_{k=1}^N \frac{(p_{A}^{jk}  + p_{B}^{jk} -2q_{jk})(V_{ij} -V_{ik})} {1+\|\mathbf{v}_j-\mathbf{v}_k\|^2}.
	\end{split}
	$}
\end{equation*}

The multiplicative update rules of the bases of both $W$ and $U$ for any $i$ and $j$ are obtained as:
\begin{equation}\label{comu}
	U_{ij}\!\! \leftarrow \!\!\frac{(A V^T)_{ij}}{(UVV^T)_{ij}} U_{ij},
\end{equation}
\begin{equation} \label{comw}
	W_{ij}\!\! \leftarrow\!\! \frac{(B V^T)_{ij}}{(WVV^T)_{ij}} W_{ij}.
\end{equation}
The update rule of the shared space preserving the coefficient matrix $V$ between the visual and semantic data spaces is:
\begin{equation}\label{comv}
	V_{ij} \!\!\leftarrow\!\! \frac{(U^TA)_{ij} \!\!+\!\! (W^TB)_{ij}\!\! + \!\!\Upsilon} {(U^TUV)_{ij} \!\!+ \!\!(W^TWV)_{ij}\!\!+\!\! \Gamma} V_{ij},
\end{equation}
where for simplicity, we let $\Upsilon=\!\! \eta\!\! \sum\limits_{k\!=\!1}^N \!\!\frac{(p_{A}^{jk} \! + \!p_{B}^{jk})V_{ik}\!+\! 2q_{jk} V_{ij}}{1\!+\!\|\mathbf{v}_j\!-\! \mathbf{v}_k\|^2}$, $\Gamma= \eta \!\!\sum\limits_{k\!=\!1}^N \!\!\frac{(p_{A}^{jk} \!+ \! p_{B}^{jk})V_{ij}\!+\! 2q_{jk} V_{ik}}{1\!+\! \|\mathbf{v}_j\!-\! \mathbf{v}_k\|^2}$.

All the elements in $U$, $W$ and $V$ can be guaranteed to be nonnegative from the allocation. \cite{lee2001algorithms} proves that the objective function is monotonically non-increasing after each update of $U$, $W$ or $V$. The proof of convergence about $U$, $W$ and $V$ is similar to that in \cite{zheng2011dimensionality,cai2011graph}.

\subsubsection{Orthogonal Projection}
After $U$, $W$ and $V$ have converged, we need two projection matrices $\mathcal{P}_A$ and $\mathcal{P}_B$ to project $A$ and $B$ into $V$. However, since our algorithm is NMF-based, a direct projection to the shared space does not exist. Inspired by \cite{cai2007spectral}, we learn two rotations to protect the data originality while projecting it into the universal space, which is known as the Orthogonal Procrustes problem \cite{schonemann1966generalized}:
\begin{equation}\label{P1}
\left\lbrace
	\begin{split}
	\mathop{\textup{min}} \limits_{\mathcal{P}_A} \|\mathcal{P}_A A-V\|, s.t. \   \mathcal{P}_A^{T}\mathcal{P}_A=I,\\
	\mathop{\textup{min}} \limits_{\mathcal{P}_B} \|\mathcal{P}_B B-V\|, s.t. \   \mathcal{P}_B^{T}\mathcal{P}_B=I,
	\end{split}
	\right.
\end{equation}
where $I$ is an identity matrix. According to \cite{zhang2015fast}, orthogonal projection has the following advantages: 1) It can preserve the data structure; 2) It can redistribute the variance more evenly, which maximally decorrelates dimensions. The optimisation is simple. We first use the singular value decomposition (SVD) algorithm to decompose the matrix: $A^{T} V = Q \mathit{\Sigma} S^{T}$. Then $\mathcal{P}_A = S \mathit{\Lambda} Q^{T}$, where $\mathit{\Lambda}$ is a connection matrix as $\mathit{\Lambda} = [I, \mathbf{0}] \in \mathbb{R}^{D\times M}$ and $\mathbf{0}$ indicates all zeros in the matrix. $\mathcal{P}_B$ is achieved in the same way. Given a new dataset $\mathcal{D}_t$, semantic embeddings $B_u=\bm{S}_u$ can be projected into $V$ as class-level UR prototypes in an unseen action gallery $\hat{V}_B=\mathcal{P}_B B_u$. A test example $\mathbf{\hat{a}}$ can be simply predicted by nearest neighbour search:
\begin{equation}\label{f1}	
\hat{y}= \underset{C+1\leqslant u \leqslant C+C_u }{\arg \max}
 \|\mathcal{P}_A\mathbf{\hat{a}}-\hat{\bm{v}}_{B_u} \|_2^2,
\end{equation}
where $\hat{\bm{v}}_{B_u}\in \hat{V}_B$. The overall Universal Representation Learning (URL) is summarised in Algorithm \ref{alg:SCPEC}.

\begin{algorithm}[htb] 
	\caption{Universal Representation Learning (URL)} 
	\label{alg:SCPEC}
	\begin{algorithmic}[1]
		\REQUIRE ~~\\ 
		Source domain $\mathcal{D}_{s}$: $A \in \mathbb{R}^{M_{1} \times N}$ and $B \in \mathbb{R}^{M_{2} \times N}$; number of bases $D$; hyper-parameter $\eta$;
		\ENSURE ~~\ 
		The basis matrices $U$, $W$, orthogonal projections $\mathcal{P}_A$ and $\mathcal{P}_B$.
		\STATE  Initialize $U$, $W$ and $V$ with uniformly distributed random values between $0$ and $1$.  
		\STATE \textbf{repeat}
		\STATE  Compute the basis matrices $U$ and $W$ and UR matrix $V$ via  Eqs. (\ref{comu}), (\ref{comw}) and (\ref{comv}), respectively;
		\STATE \textbf{until} convergence
		\STATE  SVD decomposes the matrices $A^{T}V$ and $B^{T}V$ to obtain $Q_A \mathit{\Sigma} S_A^{T}$ and $Q_B \mathit{\Sigma} S_B^{T}$
		\STATE $\mathcal{P}_A= S_A \mathit{\Omega} Q_A^{T}$; $\mathcal{P}_B= S_B \mathit{\Omega} Q_B^{T}$
	\end{algorithmic}
\end{algorithm}

\subsection{Computational Complexity Analysis}
The UAR test can be achieved by efficient NN search among a small number of prototypes. The training consists of three parts. For NMF optimisation, each iteration takes $\emph{O}(max\{M_{1}ND,M_{2}ND\})$. In comparison, the basic NMF algorithm in \cite{lee2001algorithms} applied to $A$ and $B$ separately will have complexity of $\emph{O}(M_{1}ND)$ and $\emph{O}(M_{2}ND)$ respectively. In other words, our algorithm is no more complex than the basic NMF. The second regression requires SVD decomposition which has complexity $\emph{O}(2N^{2}D)$. Therefore, the total computational complexity is: $\emph{O}(max\{M_{1}ND,M_{2}ND\}t+2N^{2}D)$, \wrt the number of iterations $t$.

\subsection{Semantic Adaptation}
Since we aim to make the UR generalise to new datasets, the domain shift between $\mathcal{D}_s$ and $\mathcal{D}_u$ is unknown. For improved performance, we can use the semantic information of the target domain to approximate the shift. The key insight is to measure new unseen class labels using our discovered `building blocks'. Because the learnt UR can reliably associate visual and semantic modalities, \ie $\hat{V}_A \sim \hat{V}_B$ we could approximate the seen-unseen discrepancy $V_A\rightarrow \hat{V}_A$ by $V_A\rightarrow \hat{V}_B$. 

To this end, we employ Transfer Joint Matching (TJM) \cite{TJM}, which achieves feature matching and instance reweighing in a unified framework. We first mix the projected semantic embeddings of unseen classes with our training samples in the UR space by $[V_A,\hat{V}_B]\in\mathbb{R}^{D\times (N_s+C_u)}$, where $V_A=\mathcal{P}_A A$. TJM can provide an adaptive matrix $\bm{A}$ and a kernel matrix $\bm{K}$:
\begin{equation}
\mathcal{L}_{TJM}(V_A,\hat{V}_B)\rightarrow (\bm{A},\bm{K}),
\end{equation}
through which we can achieve the adapted unseen class prototypes $\hat{V}_B'$ in the UR space via $\bm{Z}=\bm{A}^T\bm{K}=[V_A',\hat{V}_B']$.\\

\noindent\textbf{Unseen Action Recognition} Given a test action $\hat{\bm{x}}$, we first convert it into a kernelised representation using the trained GMIL kernel embedding in Eq. \ref{eq_nbnn_kernel}: $\bm{\hat{a}}=[\phi^1(\bm{\hat{x}}),...,\phi^{CH}(\bm{\hat{x}})]^T$. Similar to Eq. \ref{f1}, we can now make a prediction using the adapted unseen prototypes:
\begin{equation}\label{f2}	
\hat{y}= \underset{C+1\leqslant u \leqslant C+C_u }{\arg \max}
\|\mathcal{P}_A\mathbf{\hat{a}}-\hat{\bm{v}}_{B_u}' \|_2^2.
\end{equation}

\section{Experiments}
We perform the URL on the large-scale ActivityNet \cite{activityNet} dataset. Cross-dataset UAR experiments are conducted on two widely-used benchmarks, UCF101 \cite{ucf101} and HMDB51 \cite{hmdb51}. UCF101 and HMDB51 contain trimmed videos while ActivityNet contains untrimmed ones. We first compare our approach to state-of-the-art methods using either low-level or deep features. To understand the contribution of each component of our method, we also provide detailed analysis of possible alternative baselines.

\begin{table}[]
	\centering
	\resizebox{0.48\textwidth}{!}{
		\begin{tabular}{lcccc}
			\Xhline{1pt}
			Method & Feature & Setting & HMDB51   & UCF101   \\\hline
			ST \cite{2015_ICIP_Xuxun}    & BoW     & T       & 15.0$\pm$3.0 & 15.8$\pm$2.3 \\
			ESZSL \cite{2015_embarrassingly}  & FV      & I       & 18.5$\pm$2.0 & 15.0$\pm$1.3 \\
			SJE \cite{2015_akata_evaluation}   & FV      & I       & 13.3$\pm$2.4 & 9.9$\pm$1.4  \\
			MTE \cite{2016_XuXun_prioristised}   & FV      & I       & 19.7$\pm$1.6 & 15.8$\pm$1.3 \\
			ZSECOC \cite{2017_CVPR_errorCorection} & FV      & I       & 22.6$\pm$1.2 & 15.1$\pm$1.7 \\\hline
			Ours   & FV    &  I       & \textbf{24.4$\pm$1.6} & \textbf{17.5$\pm$1.6} \\\hline
			Ours   & FV    & T       & 28.9$\pm$1.2 & 20.1$\pm$1.4 \\
			Ours   & GMIL-D  & CD      & 51.8$\pm$0.7 & 42.5$\pm$0.9 \\\hline
		\end{tabular}
	}

	\caption{Comparison with state-of-the-art methods using standard low-level features. Last two sets of results are just for reference. T: transductive; I: inductive; Results are in \%.}
\label{tab_stat}
\vspace{-2ex}
\end{table}

\subsection{Settings}
\noindent\textbf{Datasets} ActivityNet\footnote{We use the latest release 1.3 of ActivityNet for our experiments} consists of $10024$ training, $4926$ validation, and $5044$ test videos from $200$ activity classes. Each class has at least $100$ videos. Since the videos are untrimmed, a large proportion of videos have a duration between $5$ and $10$ minutes. 
UCF101 is composed of realistic action videos from YouTube. It contains $13320$ video clips distributed among $101$ action classes. Each class has at least $100$ video clips and each clip lasts an average duration of $7.2$s. 
HMDB51 includes $6766$ videos of $51$ action classes extracted from a wide range of sources, such as web videos and movies. 
Each class has at least $101$ video clips and each clip lasts an average duration of $4.3$s.

\noindent\textbf{Visual and Semantic Representation}  
For all three datasets, we use a single CNN model to obtain the video features. The model is a ResNet-200 initially trained on ImageNet and fine-tuned on ActivityNet dataset. Overlapping classes between ActivityNet and UCF101 are not used during fine-tuning. We adopt the good practices from temporal segment networks (TSN) \cite{TSN2016ECCV}, which is one of the state-of-the-art action classification frameworks. We extract feature from the last average pooling layer (2048-$d$) as our frame-level representation. Note that we only use features extracted from a single RGB frame. We believe better performance could be achieved by considering motion information, e.g. features extracted from multiple RGB frames \cite{c3d2015} or consecutive optical flow \cite{twostream2014,densenet_flow_icip17,guided_flow_17}. However, our primary aim is to demonstrate the ability of universal representations. Without loss of generality, we use the widely-used skip-gram neural
network model \cite{W2V} that is trained on Google News dataset and represent each category name by
an L2-normalized 300-d word vector. For multi-word
names, we use accumulated word vectors \cite{W2V_accum}.

\noindent\textbf{Implementation Details} 
For GMIL, we estimate the pooled local NBNN kernel \cite{pooled_nbnn} using $k_{nn}=200$ to estimate the odds-ratio in \cite{local_nbnn}. The best hyper-parameter $\eta$ for URL and that in TJM are achieved through cross-validation. In order to enhance the robustness, we propose a leave-one-hop-away cross validation. Specifically, the training set of ActivityNet is evenly divided into 5 hops according to the ontological structure. In each iteration, we use 1 hop for validation while the other furthest 3 hops are used for training. Except for feature extraction, the whole experiment is conducted on a PC with an
Intel quad-core 3.4GHz CPU and 32GB memory.

\begin{table}[]
	\centering
	\resizebox{0.48\textwidth}{!}{
		\begin{tabular}{lcccc}
			\Xhline{1pt}
			Method           & Train & Test & Splits & Accuracy (\%) \\\hline
			Jain \etal \cite{2015_ICCV_ZSL_detection}      & -     & 101  & 3      & 30.3     \\
			Mettes and Snoek \cite{2017_ICCV_Spatial_aware}& -     & 101  & 3      & 32.8     \\
			Ours             & -     & 101  & 3      & \textbf{34.2}     \\\hline
			Kodirov \etal \cite{unsup_domain_Kodirov_ICCV15}   & 51    & 50   & 10     & 14.0     \\
			Liu \etal \cite{2011_Liu_zsl_action}       & 51    & 50   & 5      & 14.9     \\
			Xu \etal \cite{2016_XuXun_prioristised}        & 51    & 50   & 50     & 22.9     \\
			Li \etal \cite{domain_adapt_Li_ICIP16}       & 51    & 50   & 30     & 26.8     \\
			Mettes and Snoek \cite{2017_ICCV_Spatial_aware}& -     & 50   & 10     & 40.4     \\
			Ours             & -     & 50   & 10     & \textbf{42.5}     \\\hline
			Kodirov \etal \cite{unsup_domain_Kodirov_ICCV15}   & 81    & 20   & 10     & 22.5     \\
			Gan \etal \cite{mutlisource_zs_Gan_CVPR16}       & 81    & 20   & 10     & 31.1     \\
			Mettes and Snoek \cite{2017_ICCV_Spatial_aware}& -     & 20   & 10     & 51.2     \\
			Ours             & -     & 20   & 10     & \textbf{53.8}     \\\hline      
		\end{tabular}
	}

	\caption{Comparison with state-of-art methods on different splits using deep features.}
	\label{tab_deep}
	\vspace{-2ex}
\end{table}

\subsection{Comparison with State-of-the-art Methods}
\noindent\textbf{Comparison Using Low-level Features} Since most existing methods are based on low-level features, we observe a significant performance gap. For fair comparison, we first follow \cite{2017_CVPR_errorCorection} and conduct experiments in a conventional \textit{inductive} scenario. The seen/unseen splits for HMDB51 and UCF101 are 27/26 and 51/50, respectively. Visual features are 50688-d Fisher Vectors of improved dense trajectory \cite{idtfWang2013}, which are provided by \cite{2016_XuXun_prioristised}. Semantic features use the same Word2vec model. Without local features for each frame, our training starts from the URL. Note some methods \cite{2015_ICIP_Xuxun} are also based on a \textit{transductive} assumption. Our method can simply address such a scenario by incorporating $\hat{V}_A$ into the TJM domain adaptation. We report our results in Table \ref{tab_stat}. The accuracy is averaged over 10 random splits.

Our method outperforms all of the compared state-of-the-art methods in the same inductive scenario. Although the transductive setting to some extent violates the `unseen' action recognition constraint, the TJM domain adaptation method shows significant improvements. However, none of the compared methods are competitive to the proposed pipeline even though it is completely inductive plus cross-dataset challenge.

\noindent\textbf{Comparison Using Deep Features} In Table \ref{tab_deep}, we follow recent work \cite{2017_ICCV_Spatial_aware} which
provides the most comparisons to related zero-shot approaches.
Due to many different data splits and evaluation metrics,
the comparison is divided into the three most
common settings, \ie using the standard
supervised test splits; using 50 randomly selected actions
for testing; and using 20 actions randomly for testing.

The highlights of the comparison are summarised as follows. First, \cite{2017_ICCV_Spatial_aware} is also a deep-feature based approach, which employs a GoogLeNet network, pre-trained on a 12,988-category shuffle of ImageNet. In addition, it adopts the Faster R-CNN pre-trained on the MS-COCO dataset. Secondly, it also does not need training or fine-tuning on the test datasets. In other words, \cite{2017_ICCV_Spatial_aware} shares the same spirit to our cross-dataset scenario, but from an object detection perspective. By contrast, our CD-UAR is achieved by pure representation learning. Overall, this is a fair comparison and worthy of a thorough discussion.

Our method consistently outperforms all of the compared approaches, with minimum margins of 1.4\%, 2.1\%, and 2.6\% over \cite{2017_ICCV_Spatial_aware}, respectively. Note that, other than \cite{2015_ICCV_ZSL_detection} which is also deep-model-based, there are no other competitive results. Such a finding suggests future UAR research should focus on deep features instead. Besides visual features, we use the similar skip-gram model of Word2vec for label embeddings.Therefore, the credit of performance improvements should be given to the method itself.

\begin{table*}[]
	\centering
	\begin{tabular}{l|ll|ll|ll|ll}
		\Xhline{1pt}
		Dataset           & \multicolumn{4}{c|}{HMDB51}                                           & \multicolumn{4}{c}{UCF101}                                           \\
		Setting           & \multicolumn{2}{l}{Cross-Dataset} & \multicolumn{2}{c|}{Transductive} & \multicolumn{2}{l}{Cross-Dataset} & \multicolumn{2}{c}{Transductive} \\
		
		\hline 
		
		GMIL+ESZSL\cite{2015_embarrassingly}        & \multicolumn{2}{c|}{25.7}          & \multicolumn{2}{c|}{30.2}         & \multicolumn{2}{c|}{19.8}          & \multicolumn{2}{c}{24.9}         \\
		
		\hline
		
		UR Dimensionality & Low              & High           & Low             & High           & Low              & High           & Low             & High           \\
		
		\hline
		
		Fisher Vector     & 47.7             & 48.6           & 53.9            & 54.6           & 35.8             & 39.7           & 42.2            & 43.0           \\
		NMF (no JSD)      & 17.2             & 18.0           & 19.2            & 20.4           & 15.5             & 17.4           & 18.2            & 19.8           \\
		CCA               & 13.8             & 12.2           & 18.2            & 17.1           & 8.2              & 9.6            & 12.9            & 13.6           \\
		No TJM            & 48.9             & 50.5           & 51.8            & 53.9           & 32.5             & 36.6           & 38.1            & 38.6           \\ \hline
		Ours              & \textbf{49.6}             & \textbf{51.8 }          & \textbf{57.8}            & \textbf{58.2 }          & \textbf{36.1 }            & \textbf{42.5}           & \textbf{47.4}            & \textbf{49.9}          \\\hline
	\end{tabular}
	\vspace{1ex}
	\caption{In-depth analysis with baseline approaches. `Ours' refers to the complete pipeline with deep features, GMIL kernel embedding, URL with NMF and JSD, and TJM. (Results are in \%).}
	\label{tab_self}
	\vspace{-2ex}
\end{table*}

\begin{figure}
	\centering
	\includegraphics[width=0.22\textwidth]{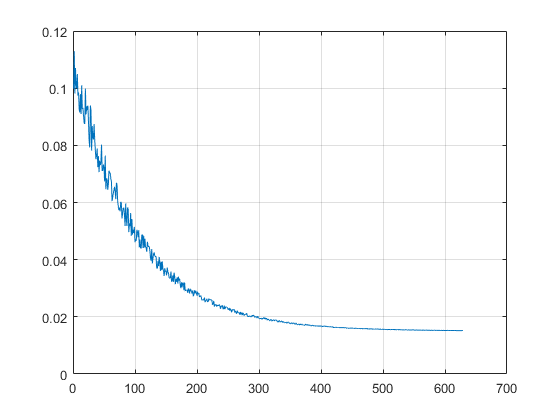}1
	\includegraphics[width=0.22\textwidth]{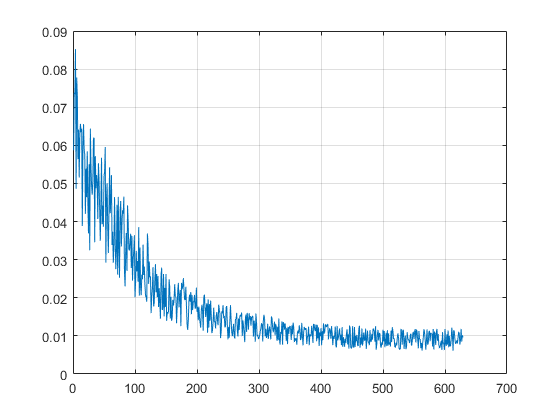}2
	\includegraphics[width=0.22\textwidth]{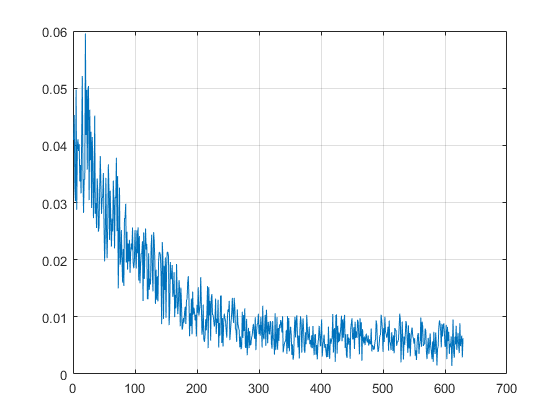}3
	\includegraphics[width=0.22\textwidth]{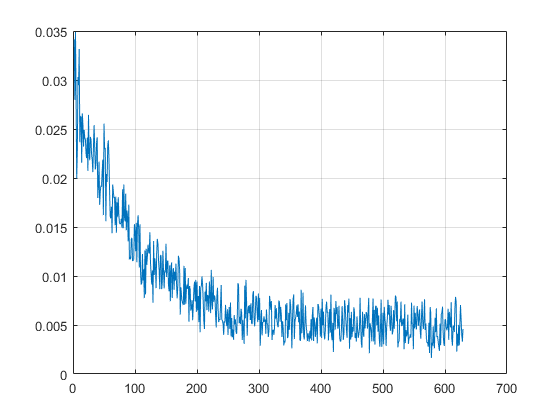}4
	\caption{Convergence analysis \wrt \# iterations. (1) is the overall loss in Eq. \ref{eq_OBJ}. (2) is the JSD loss. (3) and (4) show  decomposition losses of A and B, respectively.\label{fig_convergence}}
	\vspace{-2ex}
\end{figure}

\subsection{In-depth Analysis}
Since our method outperforms all of the compared benchmarks, to further understand the success of the method, we conduct 5 baselines as alternatives to our main approach. The results are summarised in Table \ref{tab_self}. 

\noindent\textbf{Convergence Analysis} Before analysing baselines, we first show examples of convergence curves in Fig. \ref{fig_convergence} during our URL optimisation. It can be seen the overall loss reliably converges after approximately 400 iterations. The JSD constraint in (2) gradually resolves while the decomposition losses (3) and (4) tend to be competing to each other. This can be ascribed to the difference of ranks between $A$ and $B$. While $A$ is instance-level kernelised features, $B$ is class-level Word2vec that has much lower rank than that of $A$. The alternation in each iteration reweighs $A$ and $B$ once in turn, despite the overall converged loss.

\noindent\textbf{Pipeline Validation} Due to the power of deep features demonstrated by the above comparison, an intuitive assumption is that the CD-UAR can be easily resolved by deep features. We thus use the same GMIL features followed by a state-of-the-art ESZSL \cite{2015_embarrassingly} using RBF kernels. The performance in Table \ref{tab_stat} (15.0\%) is improved to (19.8\%), which is marginal to our surprise. Such a results shows the difficulty of CD-UAR while confirms the contribution of the proposed pipeline.

\noindent\textbf{GMIL vs FV} As we stated earlier, the frame-based action features can be viewed as the GMIL problem. Therefore, we change the encoding to conventional FV and keep the rest of the pipeline. It can be seen that the average performance drop is 2\% with as high as 6.9\% in transductive scenario on UCF101.

\noindent\textbf{Separated Contribution} Our URL algorithm is arguably the main contribution in this paper. To see our progress over conventional NMF, we set $\eta=0$ to remove the JSD constraint. As shown in Table \ref{tab_self}, the performance is severely degraded. This is because NMF can only find the shared bases regardless of the data structural change. GNMF \cite{cai2011graph} may not address this problem as well (not proved) because we need to preserve the distributions of those generative bases rather than data structures. While generative bases are `building blocks' for new actions, the data structure may completely change in new datasets. However, NMF is better at preserving bases than canonical correlation analysis (CCA) which is purely based on mutual-information maximisation. Therefore, a significant performance gap can be observed between the results of CCA and NMF.

\noindent\textbf{Without Domain Adaptation} In our pipeline, TJM is used to adjust the inferred unseen prototypes from Word2vec. The key insight is to align the inferred bases to that of GMIL in the source domain that is also used to represent unseen actions. In this way, visual and semantic UR is connected by $\hat{V}_B\sim V_A\sim\hat{V}_A$. Without such a scheme, however, we observe marginal performance degradation in the  CD-UAR scenario (roughly 3\%). This is probably because ActivityNet is rich and the concepts of HMDB51 and UCF101 are not very distinctive. We further investigate the CD transductive scenario, which assumes $\hat{V}_A$ can be observed for TJM. As a result, the benefit from domain adaptation is large (roughly 5\% on HMDB51 and 1\% on UCF101 between `Ours' and `No TJM').

\noindent\textbf{Basis Space Size} We propose two sets of size according to the original sizes of $A$ and $B$ (recall section 3.2), namely the high one $D_{high}=\frac{1}{2}(M_1+M_2)$ and the low one $D_{low}=\frac{1}{4}(M_1+M_2)$. As shown in Table \ref{tab_self}, the higher dimension gives better results in most cases. Note that the performance difference is not significant. We can thus conclude that our method is not sensitive to the basis space size.

\section{Conclusion}
This paper studied a challenging Cross-Dataset Unseen Action Recognition problem. We proposed a pipeline consisting of deep feature extraction, Generative Multiple-Instance Learning, Universal Representation Learning, and Domain Adaptation. A novel URL algorithm was proposed to incorporate Non-negative Matrix Factorisation with a Jensen$\text{-}$Shannon Divergence constraint. NMF was shown to be advantageous for finding shared bases between visual and semantic spaces, while the remarkable improvement of JSD was empirically demonstrated in distributive basis preserving for unseen dataset generalisation. The resulting Universal Representation effectively generalises to unseen actions without further training or fine-tuning on the new dataset. Our experimental results exceeded that of state-of-the-art methods using both conventional and deep features. Detailed evaluation manifests that most of contribution should be credited to the URL approach.

We leave several interesting open questions. For methodology, we have not examined other variations of NMF or divergences. The GMIL problem is proposed without in-depth discussion, although a simple trial using pooled local-NBNN kernel showed promising progress. In addition, the improvement of TJM was not significant in inductive CD-UAR. A unified framework for GMIL, URL and domain adaptation could be a better solution in the future.

\noindent\textbf{Acknowledgements}
This work was supported in part by a NSF CAREER grant, No. IIS-1150115. We gratefully acknowledge the support of NVIDIA Corporation through the donation of the Titan X GPUs used in this work.

{\small
	\bibliographystyle{ieee}
	\bibliography{egbib}
}

\end{document}